\documentclass{article}
\usepackage{graphicx}

    \usepackage[preprint]{neurips_2024}
\setcitestyle{numbers,square,comma}

\usepackage[utf8]{inputenc} 
\usepackage[T1]{fontenc}    
\usepackage{hyperref}       
\usepackage{url}            
\usepackage{booktabs}       
\usepackage{amsfonts}       
\usepackage{nicefrac}       
\usepackage{microtype}      
\usepackage{xcolor}         

\title{Time-series Crime Prediction Across the United States Based on Socioeconomic and Political Factors}

\author{
  Patricia Dao\\
  \texttt{daopatricia8@gmail.com}
  \\
  \And
    Jashmitha Sappa \\
   \texttt{jashmitha.sappa@gmail.com} \\
    \And
    Saanvi Terala \\
   \texttt{saanvi.terala@gmail.com} \\
   \And
    Tyson Wong \\
   \texttt{tysonnwongg@gmail.com} \\
   \AND
    Michael Lam \\
   \texttt{michael@algoverse.us} \\
   \And
    Kevin Zhu \\
   \texttt{kevin@algoverse.us} \\
}

\begin{document}

\maketitle

\begin{abstract}
    
Traditional crime prediction techniques are slow and inefficient when generating predictions as crime increases rapidly \cite{r15}. To enhance traditional crime prediction methods, a Long Short-Term Memory and Gated Recurrent Unit model was constructed using datasets involving gender ratios, high school graduation rates, political status, unemployment rates, and median income by state over multiple years. While there may be other crime prediction tools, personalizing the model with hand picked factors allows a unique gap for the project. Producing an effective model would allow policymakers to strategically allocate specific resources and legislation in geographic areas that are impacted by crime, contributing to the criminal justice field of research \cite{r2A}. The model has an average total loss value of 70.792.30, and a average percent error of 9.74 percent, however both of these values are impacted by extreme outliers and with the correct optimization may be corrected.
\end{abstract}

\section{Introduction}
380.7 violent crimes per 100,000 people were reported in the United States by the FBI in 2022 \cite{r1A}. With high violent crime rates, it is necessary to advance crime prediction methods that aim to predict crime in different states based on social, economic, and political factors. This paper explores the use of artificial intelligence and machine learning in crime prediction using selected factors.

\section{Related Works}
 Crime predictors based on artificial intelligence and machine learning are needed to better allocate and distribute resources, like law enforcement personnel, to desired geographical areas lacking such resources \cite{r5}. Algorithms utilize datasets to predict possible crimes and recognize patterns in crime to produce predictions \cite{r17}. Indicators of crime include unemployment rates, gender ratios, high school graduation rates, and more \cite {r6}, \cite{r7}, \cite{r8}. 

\subsection{Accuracy in AI predictions}
Crime prediction is the application of mathematics to recognize any potential crime activity \cite{r26}. When generating predictions for governments, having high accuracy, the measure of a models ability to predict correctly on a dataset without bias \cite{r10} is essential. Crime forecasting: a machine learning and computer vision approach to crime prediction and prevention, a published article on the use of artificial intelligence and machine learning in crime prediction, found that out of the papers it studied, the highest accuracy was found through a Random Forest Regressor (machine learning technique for regression tasks) with a ninety-seven percent accuracy, using differing factors to predict homicides in Brazilian cities \cite{r12}, \cite{r13}. The second highest accuracy, eighty-seven percent, used a K-Nearest Neighbor (machine learning technique predicting results based on nearest neighbors) approach with chi-squared feature selection (a statistical method used to select the most relevant features from a dataset) \cite{r14}, \cite{r15}. Models with high accuracy combine artificial neural networks with machine
learning and public datasets \cite{r15}.

\subsection{Criticism in Past Techniques}
The University of Chicago developed an algorithm that predicts crime a week in advance in 2022 \cite{r16}. A separate model studied the algorithm and suggested that wealthier neighborhoods experience more arrests and crime rather than disadvantaged ones, suggesting bias \cite{r15}. 
Random Forest Regressors are high in accuracy \cite{r42}. However, it is possible for such a method to overfit, or, when an algorithm is too similar to its training data and results in an inaccurate model \cite{r33}. K-nearest neighbor approaches were second in accuracy, but K-nearest neighbors are blind and sensitive to outliers and data containing errors from the values provided \cite{r13}. High-accuracy models use artificial neural networks with supervised machine learning and public datasets, but these require a large amount of data, meaning poor amounts of data lead to poor performance in predictions \cite{r18}. Solutions to crime using artificial intelligence and machine learning often use these methods and techniques, so it is crucial to take into account such limitations to generate the most accurate prediction. 

\section{Method}
\subsection{Data Collection}
Data was gathered and cross-referenced from multiple sources to ensure accuracy for precise predictions. Each factor and dataset was taken within the timeframe 1999-2019, serving as a constant variable that increases the accuracy of the experiment. The first factor, high school graduation rates, used data from the National Center for Education Statistics \cite{r20},\cite{r100},\cite{r24},\cite{r23},\cite{r25},\cite{r26},\cite{r28},\cite{r30},\cite{r27}. Lower education levels often lead to higher crime rates\cite{r8}. The second factor, unemployment rates, came from the U.S. Bureau of Labor Statistics\cite{r30}.  This shows that financial instability often leads to crime \cite{r44}. The third factor, male-to-female percentage, was taken from KFF, with data from the Census Bureau’s American Community Survey, indicating that states with a higher male percentage may have higher crime rates due to males having higher aggression rates \cite{r37,r34}. Median income data was collected from the Federal Reserve Bank of St. Louis. The last two factors, population size, and previous violent crime rates, were taken from the Federal Bureau of Investigation. Higher population sizes often correlate with more crime due to increased interactions, and previous crime rates show law enforcement effectiveness \cite{r45}. These government-produced datasets provide high accuracy and are important for predicting violent crime.

\subsection{Model}
The model chosen was a Long-Short-Term Memory and Gated Recurrent Unit mix with lagged features, a learning rate of .001, an epoch size of one hundred, a batch size of sixty-four, rolling mean and standard deviation, with a CPU and through Google Colab. Other models were attempted, but not as accurate. As mentioned before, data was collected for the years 1999-2019, and the following features were lagged for five years previous: violent crime, population, unemployment rate, median income, high school graduation rates, political status, percent male, and the percent female, which cut the data set from twenty points per state to fifteen points per state. The numerical values of the learning rate, epoch size, and batch size were chosen based on which values would produce the best loss number, indicating higher accuracy. The rolling mean based on the previous three years, and rolling standard deviation based on the previous four years, again values chosen based on which produced the best loss number.
The model incorporated two callbacks, one of early stopping which monitored the validation loss with a patience of ten, and one of learning rate reduction which monitored the validation loss on a factor of point five and a patience of five. The model revolved around sequential data, so it was required to split the test and train datasets by the different years each data was collected. Using a time series split, the data was split into two: each state's data from 2019 for the test set, and each state's data for remaining years for the train set.

\section{Discussion}
\subsection{Results}
\begin{table}[hbt!]
  \caption{The Average Results of the First Twelve States}
  \label{sample-table}
  \centering
  \begin{tabular}{lllllllllllllll}
    \toprule
    \multicolumn{12}{c}{SA: State Abbreviation,    ADL: Average Difference Loss,     APE: Average Percent Error} \\
    \cmidrule(r){1-12}
    SA & ADL & APE & SA & ADL & APE & SA & ADL & APE & SA & ADL & APE \\
    \midrule
    AL & 1,383.8 & 5.5 & AR & -692.4 & -3.9 & CT & 1,664.5 & 3.43 & GA & 2,260.1 & 6.3   \\
    AK  & 646.0 & 10.2 & CA & 4,466.7 & 2.6 & DE & 212.9 & 5.2 & HI & 604.2 & 15.0   \\
    AZ  & 1,196.5 & 3.6 & CO & 784.3 & 3.6 & FL & 6,358.0 & 7.8 & ID & 2,056.1 & 51.4  \\
    \bottomrule
  \end{tabular}
\end{table}

To collect the results, fifty consecutive trials were ran to collect prediction data for all fifty states in the year 2019, which included the total loss, the test loss, the total execution time, and the CPU execution time.

The total loss was calculated by finding the difference of the predicted and the actual value for each state, then adding the absolute value for all of the states in a singular trial. The average total loss for the trials was 70,792.30, with a range of 34664.78, and a standard deviation of 1727.43. The test loss was found by finding the mean squared error of the predicted and the actual values. The average test loss for all of the trials was 3,708,218.88, the range is 3,341,290.88, and the standard deviation is 890,505.43.

The execution time is the total time, in seconds, it took to run each trial. The average execution time is 117.31, the range is 141.81, and the standard deviation is 47.00. The CPU execution time is the total time, in seconds, it took the CPU to run the program. The average execution time is 114.68, the range is 190.90, and the standard deviation is 55.23.

The percent error was calculated by finding the subtracting the predicted value by the actual value, divided by the actual value. The average percent error is 9.74, the range is 138.29, and the standard deviation is 21.86.

The error bar graph uses the root mean squared error to calculate the magnitude of the errors. To create the visualization with the error bar function it uses the average test loss (rmse), predictions, and actual values. Figure 1 captures various factors of variability, showcasing the accuracy of the model's prediction with prediction error and indicating the random omitting of input from using the dropout technique. Errors are assumed to be normally distributed.

\begin{figure}[hbt!]
    \centering
 \includegraphics [width=0.65\linewidth]{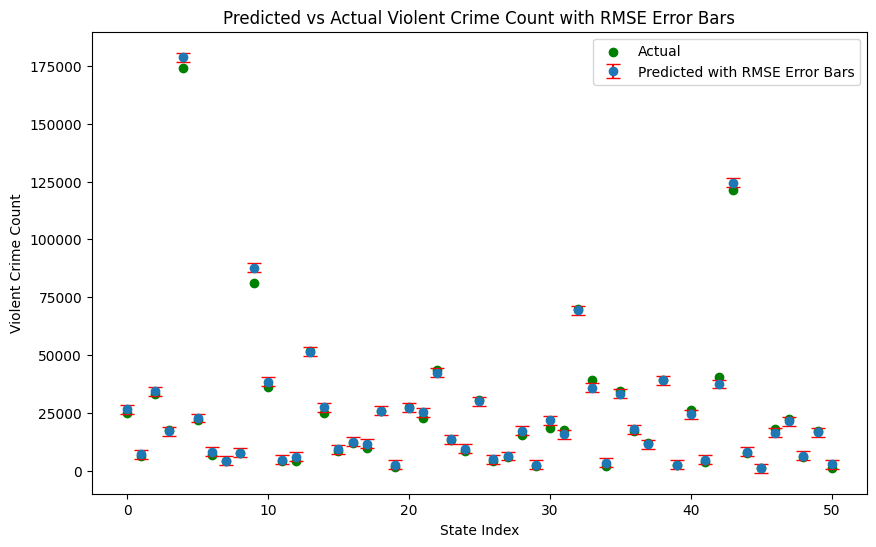}
    \caption{Error Bars Graph}
\end{figure}

\subsection{Analysis}
The relatively high number of test loss and total loss indicates that the model may have errors in compiling the best prediction of violent crime. Many of the states with a lower violent crime count contribute to these numbers much less than states with higher counts. California, with a violent crime value of 174,331, has a average loss of 4,466.73. There are exceptions to this presumption, with Florida, which has a violent crime value of 81,270, has the highest average loss of 6,358.02. However, in percent error, generally states with lower violent crime counts have the highest percent error. This is found in with Wyoming, with a violent crime count of 1,258 and a percent error of 115.20 percent.
A potential solution to reach higher accuracy is the addition of more data points and a better-tuned model to the current demands of the data.

\subsection{Assumptions}
There were assumptions made in the production of the paper and the model, the first was that all data collected was accurate and unbiased. Using an incorrect or biased dataset would lead to a biased model, which unfairly produces future predictions for violent crime.  The second assumption is that no human error occurred between finding the dataset and it appearing on the final model.
\subsection{Limitations}
There was data inconsistency between differing sources, hence why much of the data was collected from government databases, as it was deemed the most reliable source. Another was the selection of factors. Within limits of running time and unattainable data, the model did not include certain factors that could've changed the outcome of the results, such as literacy rates and mental health data. Another was the scope of the model, while it did predict on state-wide scale, initially counties were selected, however each part of data collection was done manually, which limited the size of the scale. 

\section{Conclusion}
\subsection{Potential Bias}
Bias, incomplete or incorrect data that does not accurately represent a factor \cite{r41}, may be present in datasets to be in favor of one geographical area over another in the statistics used for the model. Models and artificial intelligence may seem accurate, but the data fed to it can reflect human inequalities \cite{r41}. An example of a skewed crime predictor would be the algorithm developed by the researchers at University of Chicago as the data fed to the algorithm from the police were biased and was in favor of higher-income communities \cite{r16}. 
Existing artificial intelligence and machine learning crime predictor solutions should be tested for bias and the quality of data that is used for the algorithm to ensure the generation of accurate and trustworthy predictions \cite{r1A}. 
Going through a validation process, cross-referencing, and multi-source verification is necessary to ensure the result generates accurate predictions and not those skewed using bias.

\subsection{Going Forward}
This model indicates possible societal, political, and economic factors that increase or decrease the likelihood of crime, and with the correct efforts, can increase citizen safety and quality of life. This provides a proof of concept for future crime prediction, which is intended to be used as a point of reference in legislation for preventative measures of violent crime. With more resources in data collection and availability, it has the potential to decrease its scale to the county-wide level. Future researchers should be aware of bias in data sets and study law enforcement practices to guarantee the best predictions.

\bibliography{references1}
\bibliographystyle{abbrvnat}

\end{document}